# Clinical Document Metadata Extraction: A Scoping Review

**Kurt Miller[1,2], Qiuhao Lu[3], William Hersh[4], Kirk Roberts[3], Steven Bedrick[4], Andrew Wen[3], Hongfang Liu[3]**

[1]Bioinformatics and Computational Biology Program, University of Minnesota, Rochester, MN, USA.

[2]Center for Digital Health, Mayo Clinic, Rochester, MN, USA

[3]McWilliams School of Biomedical Informatics, University of Texas Health Science Center at Houston, Houston, TX, USA

[4]Division of Informatics, Clinical Epidemiology and Translational Data Science; Department of Medicine, Oregon Health & Science University, Portland, OR, USA

**Corresponding Author**

Hongfang Liu

McWilliams School of Biomedical Informatics

University of Texas Health Science Center at Houston

University of Texas Health

Houston, TX, USA

Hongfang.Liu@uth.tmc.edu

# Abstract

**Objectives**

Clinical document metadata, such as document type, structure, author role, medical specialty, and encounter setting, is essential for accurate interpretation of information captured in clinical documents. However, vast documentation heterogeneity and drift over time challenge harmonization of document metadata. Automated extraction methods have emerged to coalesce metadata from disparate practices into target schema. This scoping review aims to catalog research on clinical document metadata extraction, identify methodological trends and applications, and highlight gaps warranting further investigation.

**Methods**

We followed the PRISMA-ScR (Preferred Reporting Items for Systematic Reviews and Meta-Analyses Extension for Scoping Reviews) guidelines to identify articles from Ovid MEDLINE, Ovid EMBASE, Scopus, Web of Science and external sources that perform clinical document metadata extraction, either primarily as a methodology study, secondarily as a feature for a downstream application, or for analysis. We initially identified and screened 266 articles published between January 2011 and August 2025, then comprehensively reviewed 67 we deemed relevant to our study.

**Results**

Among the 67 articles included in our full text review, 45 were methodological, 17 used document metadata as features in a downstream application, and 5 analyzed document metadata composition. We observe myriad purposes for methodological study and application types. Available labelled public data remains sparse except for structural section datasets. Methods for extracting document metadata have progressed from largely rule-based and traditional machine learning with ample feature engineering to transformer-based architectures with minimal feature engineering.

**Discussion and Conclusion**

Clinical document metadata extraction research has accelerated over recent years. The emergence of large language models has enabled broader exploration of generalizability across tasks and datasets, allowing the possibility of advanced clinical text processing systems. We anticipate that research will continue to expand into richer document metadata representations and integrate further into clinical applications and workflows.

# 1. Introduction

## 1.1 Background

The digital transformation of healthcare continues to generate vast amounts of data within electronic health records (EHRs) to capture information per patient encounter [1]. Although capturing clinical information in structured formats is often preferred, these approaches suffer from limited expressiveness, rigidity, substantial design and maintenance costs, and added documentation burden for clinicians. Consequently, unstructured clinical narratives remain indispensable for representing the complexity of a patient's condition, contextual environmental and lifestyle factors, and the reasoning that informs clinical assessments and treatment decisions [2 3]. The correct interpretation of clinical information embedded in those unstructured clinical documents requires document metadata such as document type, medical specialty, author role, encounter setting, and internal section structure or layout. For example, Speier et al. [4] identify section, specialty and encounter setting information to perform topic modeling for automatic summarization of clinical reports. Sharma et al. [5] incorporate derived note section structure into a computational phenotyping system for obesity patients. More recently, retrieval augmented generation (RAG) systems integrating large language models (LLMs) with external knowledge sources, such as Nanua et al. [6] and Keerthana and Gupta [7], preprocess and index document metadata to enhance retrieval performance during inference.

However, document metadata is often incomplete, inconsistently structured, or otherwise implicit as well as varied across institutions [8]. Diverse documentation practices across departments and clinicians [9 10], migration from legacy systems [11], and drift of sublanguages and EHR use over time [12 13] contribute to extensive fragmentation of document metadata which hinder secondary use and assimilation into clinical applications. Consequently, the task of document metadata extraction or inference, as shown in Figure 1, is often required to facilitate accurate interpretation of clinical information embedded in narrative documents. Various extraction approaches such as heuristic rules, traditional machine learning methods, advanced convolutional and recurrent neural network architectures, small transformer-based models, and more recently large language models (LLMs) have been adapted to this task. While most traditional approaches may achieve adequate accuracy on certain corpora the given researchers focus on, they typically require substantial labelled data and suffer when applied to datasets from other domains or institutions. LLMs have markedly advanced performance and generalizability without labelled data required for supervised ML models or fine-tuning. However, despite recent expansions of purported context window sizes and improvements in more complex tasks such as those requiring biomedical reasoning, significant issues persist in leveraging LLMs for larger clinical applications dependent on external knowledge and/or large amounts of data. As examples, the "lost-in-the-middle problem" or "middle curse" [14 15], hallucinations and confabulations [16] [14], ambiguity of references within the input context to information outside the input context [17 18], and discordance between internal pretrained knowledge and external facts [19] [14] remain perplexing challenges across general and biomedical domains.

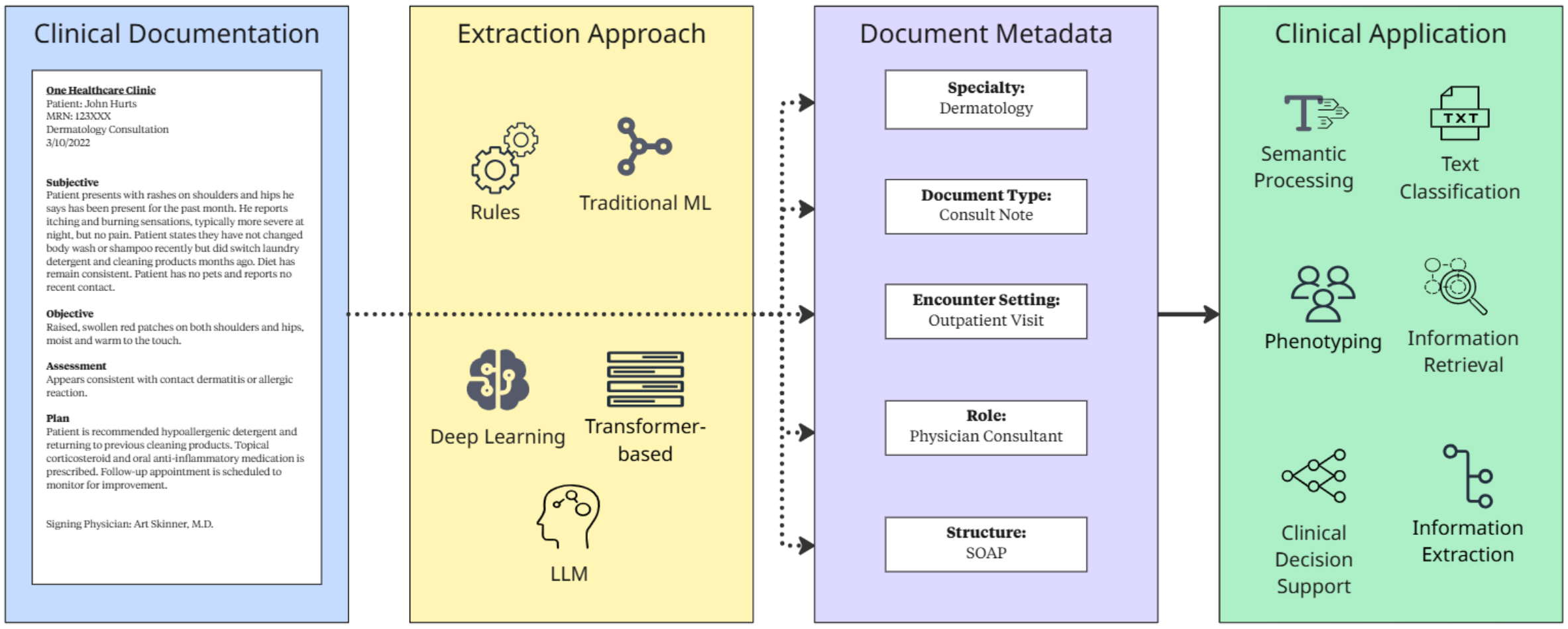

**Figure 1. Document metadata extraction and downstream clinical application workflow.** Document metadata is extracted from clinical documentation by one or multiple methods then used in a variety of possible downstream applications.

Nonetheless, supplementation of the input text in the LLM prompt with document metadata has been demonstrated to alleviate some of these issues with LLMs. For instance, performance in retrieval augmented generation (RAG) systems significantly improves when metadata is extracted prior to information retrieval during inference and included within or alongside the embedded content chunks [20] [14 21]. Similarly, other examinations of prompt engineering techniques for clinical note understanding tasks exhibited superior accuracy with shorter input contexts containing particular segments curated from the original note [22 23], suggesting that segmenting a note into its sections and including only relevant sections, or including summaries and contextual information about the input text, can yield performance improvements over the entire note text or arbitrarily chunked segments.

Despite the mounting need for this document metadata to be explicit and readily translatable between schema, limited attempts have been made recently to survey the literature for efforts to extract or analyze these document-level and structural attributes. Table 1 lists these literature reviews along with the year of the review and the dimensions of document metadata each review covers. While these investigations contain valuable compilations of clinical document metadata extraction and analysis efforts from clinical text, each review only covers certain aspects of document metadata and cover work prior to the advent of modern LLMs.

**Table 1**. Relevant Document Metadata Review Articles

| Published | Authors | Article Title | Document Metadata Reviewed |
|---|---|---|---|
| 2018 | Pomares-Quimbaya et al. | Current Approaches to Identify Sections within Clinical Narratives from Electronic Health Records: A Systematic Review | Section, Subsection |
| 2022 | Ulrich et al. | Understanding the Nature of Metadata: Systematic Review | Document Type |
| 2022 | Vorisek et al. | Fast Healthcare Interoperability Resources (FHIR) for Interoperability in Health Research: Systematic Review | Document Type, Document Structure, Role, Visit Type |
| 2024 | Peng et al. | Use of Metadata-Driven Approaches for Data Harmonization in the Medical Domain: Scoping Review | Clinical Setting, Document Type, Section |

This review aims to fill the gaps between the previous reviews by more broadly covering the literature on extracting this contextual information (i.e., document metadata) into a target schema representing an institution's internal context (i.e., schema) and by including the latest research efforts and technologies employed. By doing so, we intend to facilitate generalizable data harmonization and empower performant downstream applications which rely on contextual information about notes' content.

For this scoping review, we surveyed the literature on clinical document metadata extraction to catalog the trends of clinical document extraction data, methods, and applications. We focus on articles relating to the explicit extraction of the document and structural attributes of the document, either as a primary methodology, for incorporation in a downstream application, or for analysis. By aggregating and examining this literature, we intend to answer the following research questions:

1. What data is available and what tasks and applications do researchers leveraging that data attempt to address by using document-level metadata and note structure information?
2. What types of document attribute and structure schema are being used to capture the context of the clinical documentation?
3. What methods and architectures are employed to identify this metadata information from clinical documents?
4. What is the research landscape of clinical document metadata extraction methods and applications?

## 1.2 Working Definitions

The National Information Standards Organization (NISO) defines *metadata* as data created primarily to describe other data, differentiating types of metadata that serve to either find and understand that data more easily (i.e., descriptive metadata), relate structural elements of a resource (i.e., structural metadata), or manage data resources (i.e., administrative metadata) [24].

In the clinical domain, metadata would refer to features and elements of clinical documentation in the context of the EHR. While many definitions of general clinical metadata might include any information describing clinical data, such as report findings or whether a body of text includes specific diagnoses or phenotypes, we define *document metadata* in a clinical domain as data pertaining to the clinical context in which the document was written which helps describe and manage the document. For example, the dimensions of the LOINC Document Ontology [25] provide apt examples of descriptive document metadata. Table 2 offers the definitions of these metadata types according to the NISO [24], as well as examples of each of descriptive, administrative, and structural types of clinical document metadata in a typical EHR.

**Table 2**. Document Metadata Types and Examples in a Clinical Context

| **Metadata Type*** | **Purpose of Metadata*** | **Clinical Document Examples** |
|---|---|---|
| Descriptive Metadata | For finding or understanding a resource | Document Type (e.g., Radiology Report, Discharge Summary, Progress Note, etc), medical specialty (Internal Medicine, Neurology, etc), encounter setting (e.g., ICU, hospital, office visit, etc), personnel role (e.g., physician, nurse, PA, administrator) |
| Administrative Metadata<br>- Technical Metadata<br>- Preservation Metadata<br>- Rights Metadata | <br>- For decoding and rendering files<br>- Long-term management of files<br>- Intellectual property rights attached to content | File and data types (e.g., RTF, HTML, XML, PDF, plain text), encoding, markup language and database schema, ETL/ELT history, research authorization |
| Structural Metadata | Relationships of parts of resources to one another | Section, subsection, page number, table column headers |
| Markup languages | Integrates metadata and flags for other structural or semantic features within content | Layout, intra-document styling and formatting |

*Obtained from NISO metadata type definitions [24]

We therefore define metadata extraction from clinical documents as the process of procuring this descriptive and structural metadata from the raw document content, whether through direct extraction or by inference. We find that previous research either 1) examines an experimental methodology for extracting these types of metadata, 2) engineers the metadata as a feature for a downstream task, or 3) analyzes the composition of the metadata within or across EHRs.

### 1.3 Statement of Significance

### Problem

As the volume and complexity of EHR data increases rapidly, the need to extract and adapt contextual metadata information also grows, yet literature reviewing document metadata extraction methods and applications is sparse.

### What is Already Known

Previous surveys of document metadata extraction focus on extraction of specific document metadata attributes, such as section identification or document type classification methods, deviating definitions of metadata, or efforts toward broader data harmonization. A more comprehensive and updated compilation of the descriptive and structural metadata for clinical documentation is needed.

**What this Paper Adds**

This scoping review compiles literature exploring methods for extracting document attributes and structure, including more recent trends since the advent of transformers and LLMs. Motivations for studies investigating the viability of these methods as well as downstream applications for which these document metadata are extracted as features are also examined.

# 2. Methods

We opted for a scoping review due to the relatively underexplored nature of clinical document metadata thus far in the literature, as the volume of literature on the topic was not enough to warrant a full systematic review [26]. We leverage the Preferred Reporting Items for Systematic reviews and Meta Analyses (PRISMA) [27] extension for scoping reviews (PRISMA-ScR) [28] to perform our survey and analysis.

## 2.1 Search Strategy

For this literature review, we sought articles covering various types of clinical metadata processing published between January 2011 and June 2025. Databases searched for literature included Ovid MEDLINE In-Process & Other Non-Indexed Citations, Ovid MEDLINE, Ovid EMBASE, Scopus, and Web of Science. Our search strategy across each database was similar: the search query was composed of three distinct clauses which were each required to match at least one term. The first clause specifies some part of the medical record – “Electronic Health Record” or “EHR” or “medical record” or “clinical note” etc. The second component necessitates that a language processing task of the data occurs: “NLP” or “extraction” or “classification” or “identification”, etc. The third clause stipulates the inclusion of some kind of document metadata context: “section” or “metadata” or “document type” or “specialty”, etc.

Other articles obtained by recommendation from coauthors, web search, and via citation of papers found from the database search strategy were also included in our subsequent selection process for screening.

## 2.2 Article Selection

The search strategy yielded 225 articles total across the databases queried, while 41 additional articles were discovered through other sources. After deduplication, 160 distinct articles remained for our initial title and abstract screening. The title and abstract screening of the combined 201 abstracts from both the database query and other sources filtered out articles whose reference to contextual information and/or metadata was not the target task, part of the methodology, or subject to analysis, leaving 102 articles for subsequent full-text screening. The comprehensive full-text review used the following inclusion criteria: 1) use EHR data, 2) open access, and 4) leverage clinical document metadata (e.g., document type, section, author role, medical specialty, clinical setting, etc.) as part of either a) primary task methodology, b) secondary to a downstream task, or c) analysis. Exclusion criteria were as follows: 1) preprints, 2) closed access, 3) non-English article text, 4) only “metadata” included are domain- or disease-specific, not relating to the clinical document or encounter setting, or 5) only “contextual” information leveraged is local semantic context, as opposed to broader structural or document programmatic context. After applying these inclusion and exclusion criteria during full-text review, 65 articles were eligible for data extraction. A PRISMA-ScR flow chart of the article selection process, including categorization into the types of metadata usage, is represented in Figure 2.

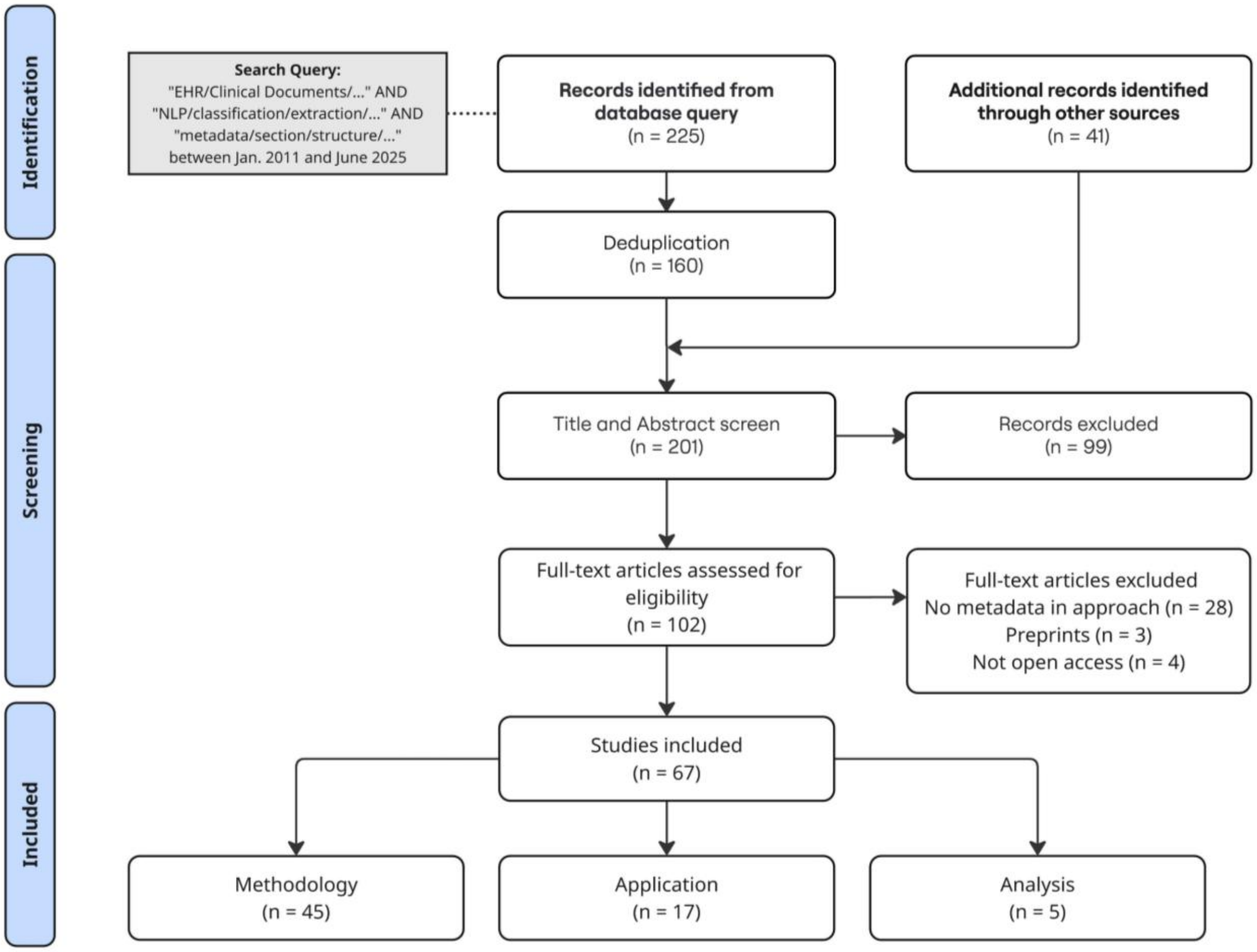

**Figure 2. Publication Selection Process.** An overview of the PRISMA-ScR approach including articles identified from database search query and additional sources, deduplicated, screened, and included in our study for full review. Reasons for exclusion of full-text articles are provided. Number of studies included for each category are also shown.

## 2.3 Data Extraction

For each of the selected articles, we collected information from the following categories:

- Dataset: the corpus or corpora used to evaluate the model, the source language of the dataset(s)
- Task: the specific task(s) facilitated by the dataset
- Metadata types: the document and/or structural metadata from the dataset incorporated into the method
- Metadata extraction approach: the method process, model architecture(s), word embedding method (if applicable), and any semantic, syntactic, and statistical representations used

- Application: if the article used the metadata extraction for a downstream application, the type of application (risk prediction, clinical decision support, phenotyping, cohort retrieval, etc.)
- Publication: the publication year, publication forum (conference or journal), and publisher subject (informatics, NLP, medical, etc)

# 3. Results

### 3.1 Data and Task

Overall, 28/67 (41.8%) publications use at least one publicly available dataset for their metadata extraction method or metadata analysis, whereas 26/67 (38.8%) of the articles included in this study leverage only a proprietary dataset and 8/67 (11.9%) use a combination of public and private datasets.

Table 3 lists the publicly available datasets containing each type of metadata and the papers which used each of those datasets for the given metadata type. The note section document structure metadata has the most public datasets available, most of which were annotated for i2b2/n2c2 shared tasks and were used for section segmentation, section classification, or both. In general, the most used publicly available datasets were corpora labelled for i2b2/n2c2 shared tasks and/or derived from the large MIMIC-III corpus.

Due to the relative dearth of publicly available datasets labelled with descriptive (non-structural) metadata, many studies relied on private, proprietary datasets annotated and adjudicated by in-house experts. A significant share of researchers tackling a section identification task also opted for labelling corpora obtained from their own institutions' EHRs, such as UCLA [4], University of Pittsburgh [29], Mayo Clinic [30], and the VA [31 32]. In these instances, the target section schema or metadata task was usually customized to a defined set of outputs unique to the study, such as a custom section schema or subset of select document types.

**Table 3**. Publicly available corpora labelled with relevant metadata

| Metadata | Public Dataset | Dataset Description | Papers |
|---|---|---|---|
| Section, Structural | 2010 i2b2/VA Challenge [33] | The "VA Challenge" dataset with manually annotated reports by i2b2 and the VA Salt Lake City Health Care System from 3 institutions on Concepts, Assertions, and Relations in Clinical Text. This dataset was subsequently annotated to include SOAP sections of clinical notes. | [34-37] |
| | i2b2 2014 shared task track 2 [38] | The i2b2/UTHealth "Risk Factors" shared task track 2, including risk factors for coronary artery disease such as diabetes, smoking, and hypertension. | [39] |
| | THYME corpus [40] | The “Temporal Histories of Your Medical Events” (THYME) dataset includes extended labels from the 2012 i2b2 temporal relation challenge dataset, containing over 1,200 consistently structured brain and colon cancer oncology notes from the Mayo Clinic. | [35-37] |
| | MIMIC-III Note Events [41] | ICU Notes from the MIMIC-III *NoteEvents* table, containing several delimited sections across a handful of different note types, can be used directly for section identification tasks as the section heading formatting and language is uniform. | [42-44] |
| | MedSecId [45] | Over 2000 MIMIC-III notes spanning 5 categories (radiology notes, consultations, echo notes, discharge summaries and physician progress notes) for 50 section types labelled to accommodate section identification tasks. | [45] |
| | n2c2 2022 shared tasks 1 and 3 [46 47] | An annotated extension of MIMIC-III Progress Notes containing SOAP labels (shared task 1) as well as problems in the Assessment sections linked to their corresponding plans for treatment in the Plan section (shared task 3) | [35 37 48] |
| | MTSamples dataset [49] | Samples of medical transcription reports across 40 medical specialties, delineated by section including SOAP sections, among additional subsections. | [50] |
| | ClinAIS at IberLEF 2023 [51 52] | Spanish clinical documents derived from the CodiEsp corpus [53] from Conference and Labs of the Evaluation Forum (CLEF) eHealth 2020 conference, made of primarily progress notes, annotated with 7 different note section types | [52 54] |
| Document Type | MIMIC-III Note Events [41] | The MIMIC-III *NoteEvents* table contains a column denoting the type of document for each note entry, enabling training and evaluation of document type classification methods. | [43] |
| | 2013 CLEF e-Health Challenge task 2 [55] | Clinical reports labelled with acronyms and abbreviations along with their normalized UMLS concept identifiers. Document headers/types are also available, allowing for differential assignment by document | [56] |
| | NHS psychiatric notes [57] | Psychiatric notes from the South London and Maudsley NHS Foundation Trust Biomedical Research Centre (SLAM BRC), harboring mental healthcare data for patients in the UK in a Case Register Interactive Search tool (CRIS) database, including different types of clinical documents | [58] |
| Specialty | iDASH 2011 [59] | Integrating data for analysis, anonymization, and sharing (iDASH) - 10 medical specialties / departments | [60 61] |
| | 2013 and 2014 Medicare Claims datasets [62] | Medicare insurance claims derived from the List of Excluded Individuals and Entities (LEIE) database from the Office of the Inspector General, available through the Centers for Medicare & Medicaid Services website (CMS.gov) | [63] |
| | Kaggle Medical Specialty Classification Dataset [64] | Patient visit notes spanning 18 medical specialties producing the document, available as a Kaggle dataset | [65] |
| | MTSamples dataset [49] | Samples of medical transcription reports across 40 medical specialties | [66 67] |
| | 2011 i2b2 VA track 1 [68] | Clinical notes and discharge summaries containing labelled coreferences of People, Problems, Treatments, and Tests. People references can refer to departments | [68] |
| | Huggingface chinese patient record triage NLP dataset [69] | 178,000 chinese patient records derived from a Medical QA dataset, containing labels for triage into 16 department classes | [70] |
| Role | 2011 i2b2 VA track 1 [68] | Clinical notes and discharge summaries containing labelled coreferences of People, Problems, Treatments, and Tests. People references can refer to personnel roles | [68] |

### 3.2 Document Metadata Extraction Approaches

Figure 3 is a histogram depicting the number of articles employing each type of modeling approach each year from 2012 to August of 2025. After the previous comprehensive review of section identification concluded in 2018 [71], the predominant method for section identification (segmentation and/or classification) and extraction of descriptive metadata has used transformer-based architectures (denoted BERT in the figure). More recently, transformers with domain-specific pretraining and LLMs have come into favor due to their relative ease of implementation, requiring no fine-tuning or less data with minimal tuning, greater generalizability across datasets, and comparable or superior performance to the previous transformer-based state-of-the-art models.

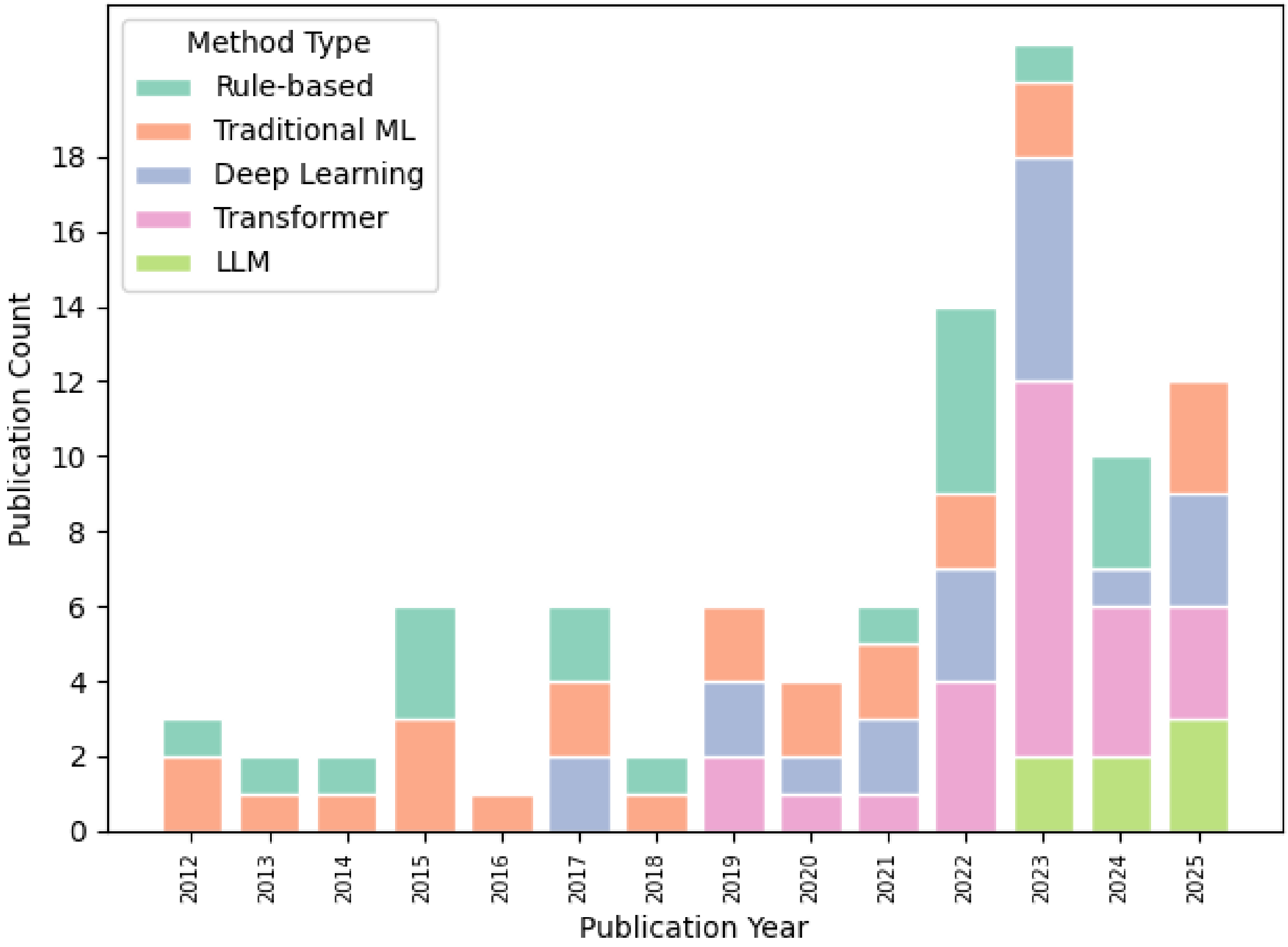

**Figure 3. Trend of method architectures used to extract clinical document metadata by publication year.** **Rule-based** approaches include logical rules, heuristics and regular expressions used to explicitly parse syntax and semantics from clinical text; **Traditional ML** refers to statistical and traditional machine-learning (ML) methods, including natural language processing (NLP)-specific algorithms that leverage machine-learning algorithms, excluding neural network derivations; **Deep Learning** includes only non-transformer-based neural network models such as multilayer perceptron, recurrent and convolutional neural network models, as well as hybrid neural network architectures; **Transformer** comprises transformer-based models with less than 7 billion parameters, whereas **LLM** refers to transformer-based large language models of equal to or greater than 7 billion parameters.

Table 4 shows the model categories, architectures, features, and ontologies leveraged across the document metadata extraction research efforts we encountered. Articles using rule-based and statistical or traditional ML as their primary, experimental models typically employed a selection from a range of possible features, techniques, and ontologies, whereas deep neural network (non-transformer) and transformer model variants relied more on embeddings. For example, Yang et al. [72] incorporated trained word2vec and skip-gram embeddings, term frequency-inverse document frequency (TF-IDF), and Medical Subject Headings (MeSH), as well as SNOMED-CT and ICD-10 codes, as features for random forest, support vector machine, gradient boosting decision tree, and logistic regression models in their section and document type classification tasks in a venous thromboembolism risk assessment application. Weng et al. [60] employed

word2vec and paragraph vector embeddings, part-of-speech (PoS) tags, TF-IDF, and UMLS and Semantic Network concept representations as features into their CNN, CNN-LSTM, and SVM models in their medical domain classification pipeline.

In contrast, more recent transformer and LLM models tested only minimal tuning or "out-of-the-box", without tuning, instead leveraging existing models pretrained on biomedical domain data. One exception to this was Zhou et al. [35], who employed continual pretraining across domain- and task-specific data to a BioBERT model to evaluate portability across disparate institutional data. None of the 6 recent implementations of LLM for metadata extraction leveraged any explicit syntactic or statistical input features, and only Socrates et al. [73] used ontological information in the form of a custom entity list for their GPT-neo experimental LLM.

**Table 1**. Model architectures and techniques

| **Category** | **Model** | **Input Features** |
|---|---|---|
| Traditional ML | Logistic Regression [5 50 72 74-76]<br><br>Decision Tree / Random Forest [5 72 76-78]<br><br>SVM [5 29 60 68 70 72 79]<br><br>CRF [32 39 56 80 81]<br><br>LDA [4]<br><br>Naïve Bayes [63 67]<br><br>Markov Model [82] | Statistical Features:<br>- n-grams [29 56 74-76]<br>- TF-IDF [60 67 72 76 77]<br>- Frequency vectors<br><br>Syntactic features (Part-of-Speech tags, orthographic features)<br>- PoS [29 60 79]<br>- Orthography [80]<br><br>Embeddings<br>- Word2vec [60 72 77 79]<br>- FastText [79]<br>- BERT [78]<br><br>Ontological Entities:<br>- UMLS [5 29 39 56 68]<br>- MeSH [68 72]<br>- RadLex [68]<br>- Semantic Network [60]<br>- SNOMED [68 72]<br>- ICD-10 [72]<br>- HPO [72]<br>- LOINC DO [81]<br>- Custom [83] |
| Deep Learning (Non-Transformer) | MLP [67]<br>CNN [60 61 65 70 79 84-87]<br><br>Sequence<br>- RNN [86-89]<br>- LSTM [65 66 77 79 84-87] | Statistical Features<br>- TFIDF [60 77 85]<br><br>Syntactic Features<br>- PoS [60 79] |
| | - GRU [86 90]<br><br>Hybrid:<br>- RCNN [91]<br>- CNN-LSTM [60 86]<br>- HGCN [89] | Embeddings<br>- Character [84]<br>- Word2vec [45 60 77 79 85 86 88]<br>- GloVe [45 65 89]<br>- FastText [45 79]<br>- FLAIR [88]<br>- Wikipedia2vec [88]<br>- Paragraph2vec [60]<br>- Custom [61 91]<br>- BERT [87] |
| Transformer-based Models | General-Purpose Models:<br>- BERT [45 65 67 81 85 86 90-92]<br>- RoBERTa [52 54]<br>- DistilBERT [67 86 87]<br>- ELECTRA [86]<br><br>Biomedical Domain Encoder Models:<br>- BioBERT [35 65 67 70 87 93]<br>- ClinicalBERT [70 76 78]<br>- BioClinicalBERT [36 44 77 87]<br>- BioMedBERT [43 87]<br>- BioMedRoBERTa [73]<br>- PubMedBERT [87]<br>- SapBERT [73]<br><br>Biomedical Domain Decoder Models:<br>- BioMedLM [36]<br><br>Multilingual and Non-english Encoder Models<br>- RoBERTa (Spanish) [54]<br>- MBERT (multilingual) [93]<br>- KMBERT (Korean) [86]<br>- GBERT/GMedBERT (German) [94]<br>- DistilBERT-pt (Portuguese) [93]<br>- BioBERTpt (Portuguese) [93]<br><br>General Domain Decoder Models:<br>- GPT-neo [73] | Methods<br>- Continual Pre-training [35 93]<br>- Active learning / weak supervision [87]<br>- Ensemble [35]<br><br>Ontological entities<br>- UMLS [87]<br>- Custom [86]<br>- LOINC DO types [81] |
| LLMs* | Biomedical Domain Decoder Models:<br>- Galactica [36]<br><br>General Domain Decoder Models:<br>- Llama 2 [37 48 95]<br>- FlanT5 XXL [36] | |
| | - ChatGPT (GPT 3.5) [37 95]<br>- Vicuna [37]<br>- Gemma 2 [48 70]<br>- Llama 3 [48 70]<br>- GPT-4 [37 95 96]<br>- Mistral [48]<br>- GLM [70]<br>- Qwen [70]<br><br>Fine-tuned/Instruction-tuned Models<br>- Tulu2 [37]<br>- Mistral OpenOrca [48] | |

*LLMs are defined as transformer-based models with >= 7 billion parameters

### 3.3 Applications

Among the 62 articles we consider in our review that perform clinical document metadata extraction (i.e., excluding "analysis" papers), 17 (28.6%) identify the document metadata in a preprocessing step to use as a feature in a downstream application. In our review, refer to these as "application" papers. Across these 17 articles, we found a variety of specific purposes for which document attributes were extracted and used as features. For this review, we group these purposes into six broader application type categories: text classification, information extraction, information retrieval, semantic processing, computational phenotyping and clinical decision support. Figure 4 is a Sankey diagram illustrating the use of public or proprietary data (stage 1); document metadata component(s) extracted (stage 2); whether the purpose of the article was a methodology study, application, or analysis (stage 3); and, for the articles that target a downstream application, the application category (stage 4).

Of the 17 application papers, 9 (52.9%) incorporate a public dataset. Regarding the types of document attributes, 14/17 (82.4%) used section information and 4/17 (23.5%) used document type, whereas 7/17 (41.2%) incorporated more than one attribute as input features to a model performing a different, downstream task. One example of an application of multiple attributes was the work of Speier et al [4], who used document type, physician medical specialty, and relative document dates for their target tasks of computational phenotyping and text classification. Kugic et al [95] integrated structural (section) and medical specialty information into their pipeline for biomedical and clinical acronym disambiguation and resolution in English, German, and Portuguese clinical texts.

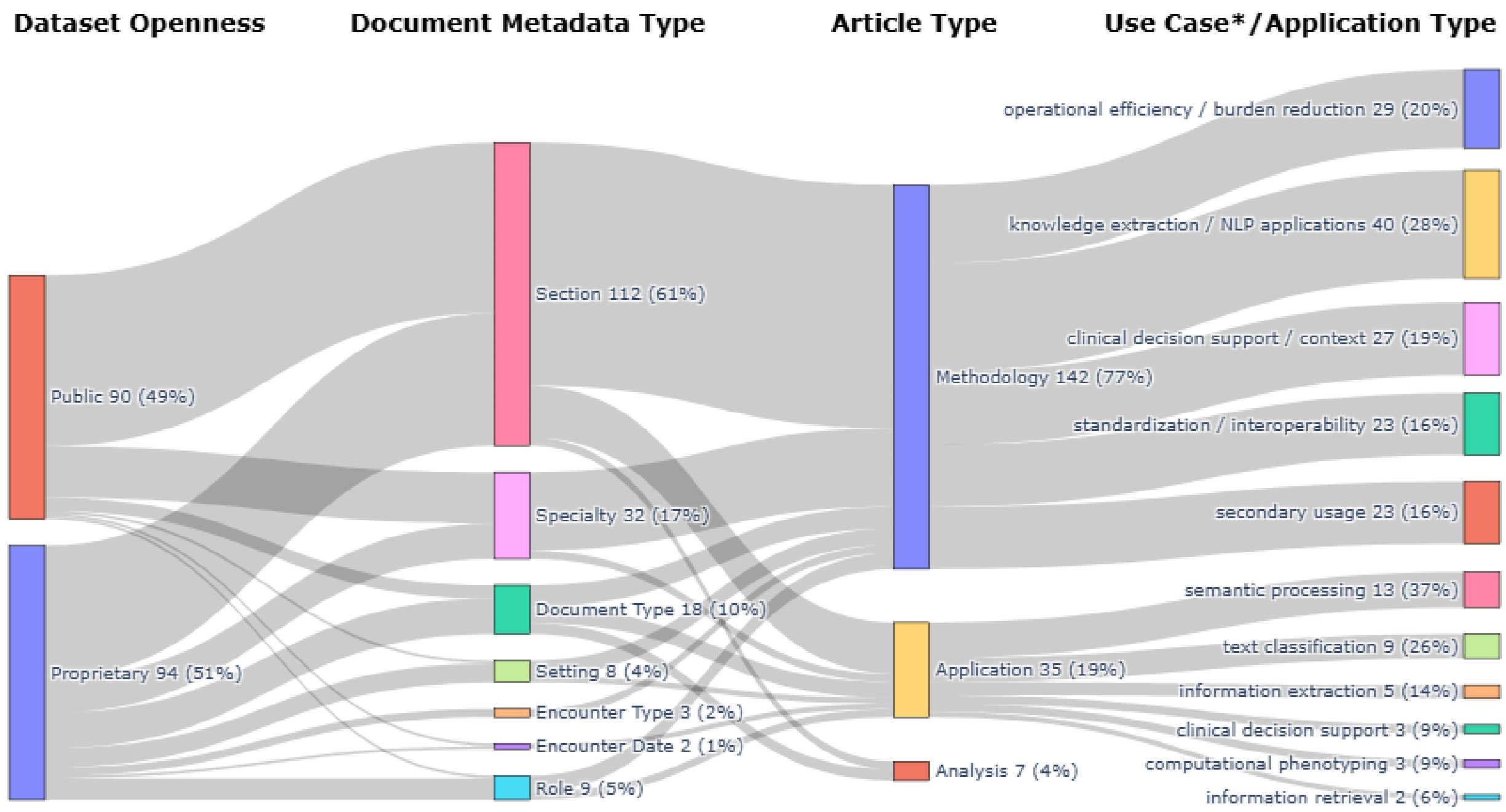

**Figure 4. Clinical document metadata datasets, metadata types, article type, and applications.** Counts next to each node title represent the unique combination of the number of corpora in the reviewed articles falling under that category and number of methodology motivations: a single article may be represented multiple times in this Sankey diagram if it included multiple datasets for model training and evaluation and/or is a methodology study including multiple motivations for conducting the study. The percentage of that count within the tier is also shown. The first tier *Dataset Openness* represents whether the dataset is publicly available or private/proprietary. The second tier for *Document Metadata Type* represents the document attribute or structure dimension labelled in each dataset. The third tier *Article Type* represents whether the dataset was used to evaluate a methodology (i.e., a methodology study), included as a feature for a downstream application, or a subject of analysis. The fourth tier represents either which use case(s) methodology studies mention as a motivation or for which downstream application type an application article uses the document metadata.
*Use case (methodology motivation) categories were created using generative AI.

Between the six categories of application that we observed, computational phenotyping and semantic processing tasks such as disease classification and coreference resolution, respectively, were the most common. However, the distribution of these applications was distributed relatively evenly across the categories.

### 3.4 Research/publication context

Figure 5 displays the frequency of publication venue types by research topic. Informatics journals and conferences were the most common venues for clinical document metadata research, representing 25/67 (37.3%) and 16/67 (23.9%) of the publications, respectively, or 41/67 (61.2%) of the publications overall included in our review. Computer science/AI/NLP venues were also a relatively common destination, representing 20/67 (29.9%) of the publications included here. Although medical/health/bioscience venues were considerably less common overall, their frequency appears to be increasing since the advent of larger language models in late 2022 and 2023.

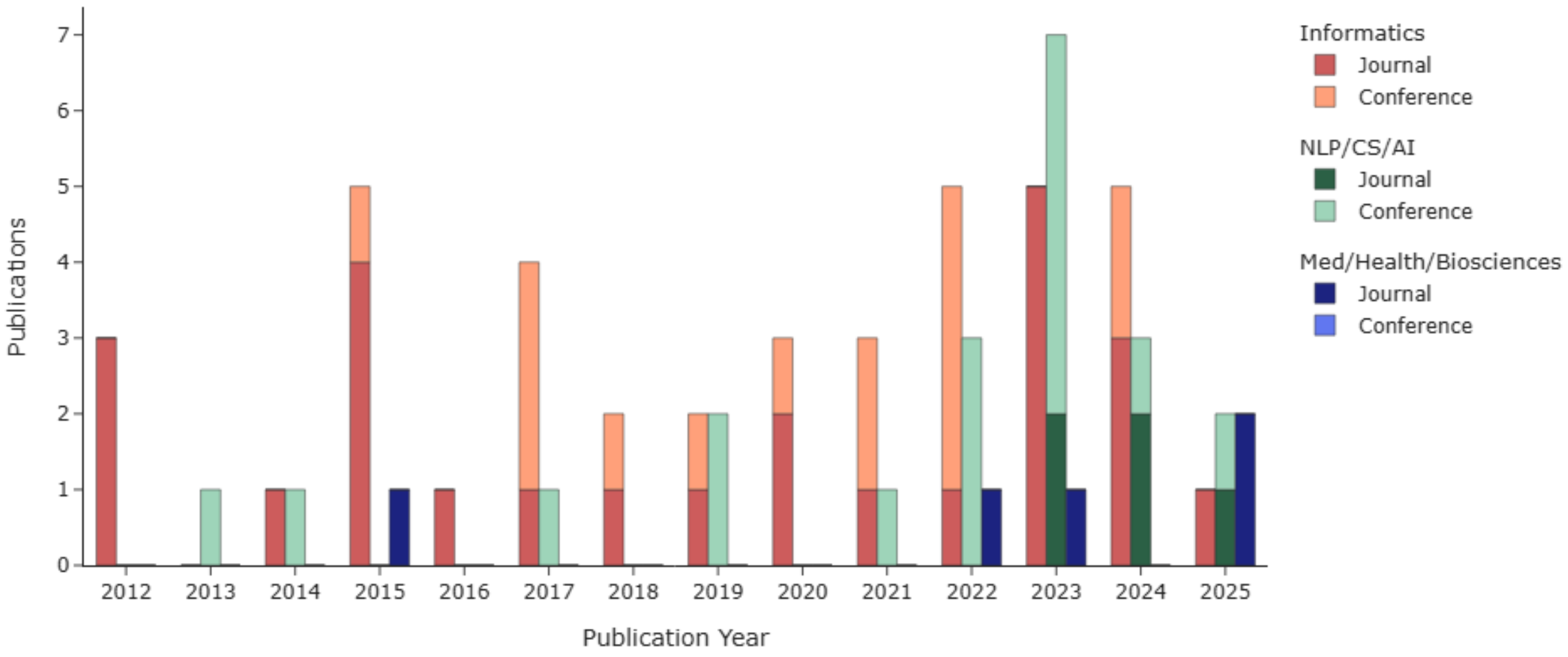

**Figure 5. Publication Venue Type and Topic Counts.** The journal and conference publications of each venue type are stacked for a given year. Publications in 2025 include through the month of August.

# 4. Discussion

For this scoping review, we have surveyed the data, methods, and applications that incorporate document metadata extraction, either as an exploration of a methodology for a primary task, as a feature for a downstream clinical application, or that conduct metadata analysis. Our findings point to a few general patterns and more recent trends warranting further consideration: the dearth of public document metadata corpora, the evolution of contextual modeling, the diversity of document metadata schemas, and the fragmentation of the publication landscape. Here, we expound on those patterns, then discuss the current challenges and future frontiers as well as the limitations in conducting this review.

## 4.1 Dearth of Public Clinical Document Attribute Corpora

While we did find several public corpora containing note section labels and the SOAP schema was a somewhat commonly recurring standard, relatively few used the same set of sections. Moreover, corpora with annotated descriptive metadata were relatively sparse, as document type and medical specialty only had 3 and 4 public datasets, respectively. Although sometimes the information can be inferred or derived from other sources, public data was even more sparse for clinical setting and author role, as we found no datasets which explicitly labelled and advertised these document features. As a result, the lack of publicly available data, along with the high cost of corpus annotation, could be inhibiting further clinical document metadata extraction research.

Except for the document type and section label data that could be juxtaposed from MIMIC-III data and the specialty and document type data available for the MTsamples transcript dataset [49], none of the other publicly available datasets comprise multiple different types of document metadata, though some datasets other document structure or attribute information could be inferred from the content, such as simple section identification rules for the CLEF eHealth Challenge 2013 task 2 dataset [55].

### 4.2 Evolution of Contextual Modeling

Since the emergence of the EHR as a primary source of patient information, the contextual modeling approaches of document metadata appear to have undergone three somewhat distinct phases of development. The first phase, lasting until roughly 2015, relied heavily on hardcoded rules and regular expressions to identify key delimiters and terms within the text and targeted section identification tasks. These were almost exclusive methodological studies or applications of semantic processing. Mowery et al. [29], Dai et al. [80], and Ganesan et al.[74] exemplify this phase in their studies evaluating rule-based section segmentation and classification methods. The second phase spanned roughly 2015 until late 2023, when traditional ML, more advanced NLP, neural network and transformer-based architectures were predominantly leveraged in tandem with an assortment of feature types and feature engineering processes. Embeddings, syntactic and statistical features, and semantic or ontological representations, for example, were tested during this era of contextual modeling. For transformer models such as BERT, additional pretraining on biomedical domain corpora such as for BioBERT, PubMedBERT and BioClinicalBERT, as well as additional parameter tuning such as fine-tuning and continual pretraining, were also employed and evaluated as potential avenues to improve performance and generalizability. The most recent phase, starting roughly around the advent of ChatGPT in early 2023, relaxed the demand for feature engineering and expanded the dataset and task variety in each study.

The progression through these phases has coincided with increased capability of model types and latent embedding spaces to represent words in their context, from explicit to implicit and from monosemy to polysemy and homonymy. Rule-based and statistical ML modeling from the first phase of the review timeframe entails explicit dictionaries mapping words and/or concepts to inferences, such as document scores for TF-IDF. Early word embedding models such as word2vec use implicit relative representations of points in a high-dimensional vector space yet only include only a single representation for each word. Later embeddings used in deep learning architectures (e.g., CNNs, LSTMs, and transformers), such as GloVe and BERT contextual embeddings, integrated many definitions for a single word based on their context (i.e., polysemy) including words that share the same spelling but may have starkly contrasted meanings based on their context (i.e., homonymy). These contextual embeddings, learned during the pretraining and reinforcement learning from human feedback (RLHF) phases of model development, implicitly impart context-based semantics thereby empowering greater generalizability across disparate semantic contexts. The advancement from explicit to implicit modeling and monosemy to polysemy and homonymy has spurred recent interest in generalizability across tasks, transferability across domains, and integration into larger, more complex applications.

### 4.3 Diversity of Clinical Document Metadata Schemas

The sparsity of publicly available corpora with annotated document-level metadata as well as a lack of standardization across institution documentation and document metadata practices has necessitated a reliance on proprietary datasets bearing many custom schemas tailored to each study's unique research interests. While some section identification attempts still target the conventional SOAP and CDA schema, a greater proportion of these either used a modified version of these schemas or delineate their own set of sections. Other document attributes such as medical specialty and document type may include only a subset

Due to the heterogeneity of schemas across healthcare institutions, there is a demand for flexible solutions that can take a source dataset of any metadata schema and adapt it to a standardized or target schema. Recognizing this problem, Peng et al. [97] recently investigated efforts toward data harmonization using ETL/ELT systems that can translate metadata between different institution's schemas.

Recent advances in model architectures and techniques allow internal encoding of knowledge about general language imparted during pre-training (e.g., masking on general domain corpora) and post-training (e.g., RLHF) phases have enabled study of generalizable, portable solutions that bridge these heterogenous metadata schemas.

### 4.4 Diverging Definitions of Metadata

Despite the NISO's unambiguous definition of metadata and delineation of its subtypes [24], the systematic review by Ulrich et al [98] on the nature of metadata reveals a discordance among researchers as to what constitutes metadata and how to taxonomize it: hard and soft metadata [99], record-level and data value-level metadata [100], and context-dependent and context-independent metadata [101]. Our observations on the usage of the term "metadata" as well as on the diversity of document metadata schemas in the literature we reviewed corroborate those findings. For example, Kim et al [102] define a metadata ontology relating clinical concepts identified from the document into a formal tree structure, and Caufield et al [103] propose a clinical case report extraction approach for obtaining metadata as summaries of medical content contained within the given report. These metadata were excluded from our review because they do not conform to our definition of metadata as document-level and structural dimensions of metadata, but rather whether they include particular semantic information related to each of their respective application tasks.

As "metadata" is generally more loosely defined as *any* data about data and the amount of ways data can be described also grows, the amount of metadata schema and definitions of "metadata" will also continue to grow. This rapid expansion of data descriptions, structures, and types underscores an emerging need for a generalizable way to harmonize data from any source to a target EHR metadata schema.

### 4.5 Manifold Methodological Motivations and Downstream Applications

The final (rightmost) tier in Figure 4 highlights the distribution of downstream applications for which document metadata extraction methodologies are used. These categories are high-level abstractions aggregated from our observations—the landscape of these motivations and applications was considerably diverse, as each category depicted in the diagram represents a range of specific use cases. For example, motivations for studies of document metadata extraction methodologies include reducing administrative and physician burden by navigating insurance claims and suggesting appropriate tags using section information such as the work of Nair et al. [104], improving department triage using medical specialty prediction such as in Lee et al. [86], as well as improving downstream applications like cohort discovery such as our work in Miller et al. [48]. For applications, "semantic processing" represents a group of specific use cases pertaining to identifying and relating semantic components of text, including named entity recognition, coreference resolution, relation extraction, and acronym resolution, among others.

Further, the opportunity provided by LLMs will likely facilitate the application of document metadata to more applications not detected as published, peer-reviewed and open-access work for this scoping review, yet we observed are on the horizon. The myriad motivations and applications for which we found extraction of document metadata useful demonstrate the breadth of use cases to which it can be applied across healthcare and research.

### 4.6 Mounting Interest in Document Metadata

While some document attribute and structure extraction has been a topic of research prior to and through the beginning of the scope of this review, we observe a trend of increasing research over the review period. The increase in journal and conference publications in the latter half of our review period, from 2018 to 2025, in Figure 5 illustrates this growth. Figure 5 also exhibits a greater diversity of publication venues in the latter half of the study period, showing that more journals and conferences accepting articles analyzing document metadata composition or investigating extraction methods. Moreover, the growth in document metadata interest does not appear to be limited to the English language. Although sparse non-English research existed prior to 2018, since then document metadata extraction methods have been explored across Spanish, Portuguese, German, Mandarin, Korean, and Finnish corpora. These trends of increasing journal and conference publications, diversity of publication venues, and breadth of natural language corpora exhibit growing global interest in document metadata research.

### 4.7 Current Challenges and Future Direction

Clinical applications leveraging large documents such as those employing information retrieval have recently shown performance improvement when incorporating metadata filtering prior to dense or hybrid embedding search [14] or leveraging document metadata from the whole document or adjacent text segments to the target segment during indexing [21 48]. However, in information retrieval, challenges remain for how best to chunk information from these large documents, what information to include with the chunk when indexing, how to transform a query to precisely target the appropriate information, and how to retrieve and rank results using such document metadata. Which document metadata and other contextual information to introduce the model to in the prompt and in which stages will be an ongoing thread of exploration.

Likewise, structural metadata is more than the section composition and order highlighted in this review. As the NISO definitions of the types of metadata distinguishes, structural metadata can also be visual layout or other information captured in html/xml or markdown, such as relative positioning, tabular information such as column headers, use of bullet points or numbering, and formatting/style which may communicate emphasis or otherwise provide context surrounding given text. Capturing this information is a challenging ask for the scanned images and PDF documents transferred between institutions as part of outside medical records. This challenge is discussed somewhat extensively by Goodrum et al. in [76], but otherwise there is relatively little examination of this emerging challenge across clinical informatics research, possibly because multimodal research integrating visual layout and textual information is still relatively young. Broader research efforts in "layout parsing" of unstructured documents with visual content such as PDFs into semi-structured markup (e.g., html) or markdown has been ongoing, but recent advancements in modeling using LLMs, such as SmolDocling [105], have recently made rapid progress in state-of-the-art performance on related tasks.

### 4.8 Limitations

The selection of a few document-level and structural attributes among the diverging definitions of clinical document metadata, perhaps bias our review toward those document-level and structural attributes we selected for this article. Descriptive metadata about alternative knowledge sources such as patient data or medical research, summary metadata about impressions or findings extracted from the document, type of procedure or imaging performed, and the body location of an imaging report could be considered as metadata at the document level yet were omitted from our search to constrain the scope of the review. Temporal information about the document or clinical event (e.g., document signed date or date of procedure) could be also considered a dimension of document metadata but was not covered since temporal modeling has its own expansive field of literature that would be infeasible to include here.

Document-level and structural metadata extraction methods, especially those leveraged for downstream applications, were commonly absent from abstracts and buried in descriptions of methods, sometimes tersely depicted or even omitted from methods descriptions entirely. This obscurity was especially prevalent in studies that leverage metadata for a downstream application, since the application task was often the focus of the article instead of the feature engineering process for the constituent metadata. Further, conflation of the aforementioned local, linguistic context methods with the type of programmatic context we focus on in this study exacerbated the challenge of finding articles that meet our specific search criteria since mentions of “contextual information” often referred to only the local semantic context. Thus, due to the difficulty in searching for and screening these methods including a programmatic context component, we may have missed relevant articles or biased our search toward articles that referenced document metadata in particular ways.

# 5. Conclusion

Clinical document metadata is crucial for proper comprehension of patient encounter information and facilitating downstream applications, especially as unstructured clinical text captured about patient encounters becomes increasingly vast and heterogenous across healthcare organizations. We conducted a scoping review of clinical document metadata extraction research from 2011 to 2025 and characterized the data, methods, applications, and research context of the 67 articles we found that represent the literature. We found that while publicly available data is limited, methods and applications are advancing, interest in document metadata is growing, and the publication landscape is increasingly fragmented. We suspect that these trends will continue as document metadata and its associated data content continue to evolve and document metadata remains crucial for contextualizing unstructured clinical text content.

# Declaration of competing interests

The authors declare that they have no competing interests.

## CRediT Authorship Statement

**Kurt Miller:** Writing – original draft, Methodology, Data curation, Investigation, Formal Analysis, Conceptualization, Visualization, Validation. **Qiuhao Lu:** Investigation, Data curation. **William Hersh:** Writing – review and editing, Conceptualization, Methodology. **Kirk Roberts:** Writing – review and editing, Conceptualization, Methodology. **Steven Bedrick:** Writing – review and editing, Conceptualization. **Andrew Wen:** Writing – review and editing, Investigation. **Hongfang Liu:** Writing – review and editing, Conceptualization, Methodology, Investigation, Supervision, Project administration, Resources.

## Funding

This work was supported by the National Institutes of Health grant numbers R01LM011934 and R01LM014508.

## Acknowledgements

We thank Larry Prokop, Librarian, in the Mayo Clinic Libraries for contributions in drafting the search queries and conducting the search of databases for articles reviewed in this study.

## Data Availability

See Supplementary Material.